\documentclass{article} 
\usepackage{colm2024_conference}

\usepackage{microtype}
\usepackage{hyperref}
\usepackage{url}
\usepackage{booktabs}
\usepackage{amsmath}
\usepackage{graphicx}
\usepackage{placeins} 
\usepackage{subcaption}
\usepackage[toc,page]{appendix}
\definecolor{darkblue}{rgb}{0, 0, 0.5}
\hypersetup{colorlinks=true, 
citecolor=darkblue, 
linkcolor=darkblue, 
urlcolor=darkblue}
\usepackage[bottom]{footmisc}
\title{\textsc{Cognitive Phantoms in LLMs Through the Lens of Latent Variables}}

\author{Sanne Peereboom$^1$, Inga Schwabe$^1$ \& Bennett Kleinberg$^{1, 2}$  \\
\quad $^1$Department of Methodology and Statistics\\
\quad \quad Tilburg University\\
\quad \quad Tilburg, the Netherlands \\
\quad $^2$Department of Security and Crime Science\\ 
\quad \quad University College London\\
\quad \quad London, UK\\
\texttt{\{s.peereboom, i.schwabe, bennett.kleinberg\}@tilburguniversity.edu} \\
}

   \colmfinalcopy 
 
\begin{document}

\maketitle

\begin{abstract}
Large language models (LLMs) increasingly reach real-world applications, necessitating a better understanding of their behaviour. Their size and complexity complicate traditional assessment methods, causing the emergence of alternative approaches inspired by the field of psychology. Recent studies administering psychometric questionnaires to LLMs report human-like traits in LLMs, potentially influencing LLM behaviour. However, this approach suffers from a validity problem: it presupposes that these traits exist in LLMs and that they are measurable with tools designed for humans. Typical procedures rarely acknowledge the validity problem in LLMs, comparing and interpreting average LLM scores. This study investigates this problem by comparing latent structures of personality between humans and three LLMs using two validated personality questionnaires. Findings suggest that questionnaires designed for humans do not validly measure similar constructs in LLMs, and that these constructs may not exist in LLMs at all, highlighting the need for psychometric analyses of LLM responses to avoid chasing cognitive phantoms.

$\\$
\textit{Keywords: large language models, psychometrics, machine behaviour, latent variable modeling, validity}
\end{abstract}

\section{Introduction}
Large language models (LLMs) are becoming progressively intertwined with day-to-day life. LLMs are commonly used via easy-to-access interfaces such as ChatGPT to retrieve information, obtain assistance for homework, provide customer service, and so on. With an increasing number of parameters and more training data, LLMs become capable of processing and generating nuanced natural language \cite{brown2020}. For example, there is evidence that LLMs have generated text that was perceived as human more often than comparable human-written text \cite{jakesch2023} and that they are capable of advanced reasoning tactics and negotiation, ranking among the highest-level players in a strategy game with human players \cite{meta2022}. 

The continuing evolution of LLM capabilities comes with the need to understand the models better, and increasing body of work has started to study LLMs with regard to their behaviour. For example, social biases present in training data were found to become ingrained in word embeddings \cite{caliskan2017}, and LLMs have occasionally been found to otherwise misalign with human values \cite{kaddour2023}: a challenge that is yet unsolved and carries implications for AI safety. There are efforts to better understand the origins and solutions to such behaviours, however, a better understanding of LLMs comes with a significant challenge: the billions of parameters contained in the models significantly complicates the analytic assessment of the models' inner workings (e.g., extensive adversarial testing is a common approach to detecting potential vulnerabilities that could result in harmful LLM responses \cite{mozes2023}). In other words, alternative approaches to studying LLM behaviour are needed.


\subsection{Machine behaviour and machine psychology}
As an alternative to analytic assessment of the underlying processes in LLMs, adopting a \textit{machine behaviourist} perspective can be useful \cite{rahwan2019}. Inspired by the study of animal behaviour, machine behaviour is the study of the behaviour manifested in intelligent machines in terms of development, evolution, function, and underlying mechanisms \cite{rahwan2019}. Building further on parallels between the study of the human mind and intelligent machines, \textit{machine psychology} \cite{hagendorff2023} refers to the evaluation of LLMs analogous to participants in language-based psychological studies.

Early findings in the field of machine psychology suggest a semblance of humanness in LLMs. For example, the GPT-3 model was found to be prone to human-like cognitive errors in classic psychological decision-making tests \cite{binz2023}. Other studies have used existing questionnaires to measure personality traits in LLMs \cite{miotto2022, huang2023} and psychological profiles of LLMs at a larger scale \cite{pellert2024}, one study even reporting a high degree of dark personality traits (psychopathy, Machiavellianism, and narcissism) in GPT models compared to average scores in a human sample \cite{li2024}.

The majority of aforementioned constructs would be considered \textit{latent variables} in psychological theory: these constructs are not directly observable nor directly measurable. Instead, these variables are \textit{indirectly} measured through measurable behaviours hypothesised to be caused by the underlying latent trait. This approach can be a powerful supplement to the analytic assessment of LLMs: identifying overarching latent phenomena that cause a tendency towards undesired responses (e.g., dark personality patterns that may increase the risk of toxic responses) could help guide targeted adversarial testing strategies for more efficient prevention of harmful output. The indirect measurement of these latent phenomena, usually through a questionnaire or test, has a longstanding tradition in quantitative psychology and is the foundation of the discipline of psychometrics.

\subsection{Latent variables and psychometrics for LLMs} 
Administering readily available questionnaires seems like a quick way to accurately measure latent traits in LLMs. Many studies on latent traits in LLMs apply existing measurement instruments, constructed for human samples. In psychometric terms, these studies thus aim to infer from LLMs various latent traits that might systematically affect model behaviour. A test or questionnaire is administered to an LLM, and the resulting responses are aggregated into composite scores (e.g., a mean dimension score) and interpreted - often in relation to human samples - as a proxy for the latent trait of interest. However, this approach relies on crucial yet rarely acknowledged assumptions regarding the \textit{validity} of the measurements in LLMs. 

The validity of a questionnaire refers to the extent to which it measures what it is intended to measure. It depends firstly on the existence of the latent trait, and secondly on the notion that changes in the latent trait cause changes in the observed responses \cite{borsboom2004}. Administering an existing questionnaire validated on human samples presupposes that the latent trait of interest exists in LLMs. It further presupposes that the administered questionnaire measures the respective trait, and that it is measured equivalently in humans and LLMs. A questionnaire that is validated on human samples certainly does not guarantee that it is valid for LLMs: LLMs will always generate \textit{some} response, which may be used to calculate a mean score on certain latent dimensions (e.g., the degree of psychopathy in the models), but that does not mean that this score is meaningful or that the latent phenomenon exists in LLMs at all. Relying on composite scores without acknowledging this problem may thus create the illusion of humanness in LLMs.

Although a handful of studies have performed psychometric evaluations of LLM responses, assessments have either been limited to reliability or validation studies using composite scores \cite{huang2023b, serapio-garcia2023}, or estimating reliability without considering validity \footnote{A reliable questionnaire is not necessarily valid. Reliability reflects the precision with which a questionnaire measures \textit{some} trait \cite{mellenbergh1996}, whatever it may be. Validity should precede reliability: the precision of a questionnaire is not very meaningful without (1) the reasonable belief that the latent trait of interest exists in the first place, and (2) the reasonable belief that the questionnaire measures this trait, and not some other trait.} \cite{shu2023}. Others acknowledge the need to evaluate whether tests correctly measure a latent trait of interest in LLMs \cite{pellert2024, wang2023}, although this notion still presupposes that the latent trait exists in LLMs to begin with. 

\subsection{The validity problem for LLMs}
Validity is not a methodological problem but rather a theoretical one \cite{borsboom2004}. The existence of latent phenomena and their causal effects on the measurement outcomes cannot be proven due to their unobservable nature - they must be founded in extensive psychological theory to say that there is a justifiable expectation that a test or questionnaire reasonably measures the phenomenon it purports to measure \cite{borsboom2004}. This is already a difficult problem in human psychological research, let alone for machine psychological research where substantive theory does not yet exist. 

Although validity cannot be established through empirical methods, validation studies can supplement theory-based hypotheses on latent phenomena and their causal effects on measurable behavioural outcomes. For example, we can test whether there is a common latent factor that causes the covariance among a set of items that should reflect the same underlying trait. This approach does not guarantee that the common factor represents a \textit{specific} latent phenomenon, nor does it provide information on the actual causal process at play between a latent phenomenon and measurement outcomes. However, it does provide plausibility for the measurement model under the assumption that the latent phenomenon exists and that it can reasonably be measured using the measurement instrument.

An additional complexity for LLMs pertains to a measurement unit problem, namely, whether an LLM can be considered analogous to a "population" versus an "individual". This issue is straightforward for humans but unknown for LLMs, yet imperative to the application of appropriate methods. As LLMs are trained on vast corpora of human-written data, they represent a diverse range of information, perspectives, and writing styles extracted from text written by countless individual people. When prompted, the LLM samples from its learned distribution to generate a response. An LLM could be seen as analogous to an approximation of the population distribution of language and information in the training data, or as analogous to a single "average" individual. This distinction has far-reaching implications for validation studies, because no instrument can be validated on a single individual.

\subsection{The present study}
While approaches from machine behaviour and psychology can be a useful guiding framework for studying LLMs (and AI models more generally), we argue that the current underlying measurement theoretical approach is insufficient to fully grasp the nuances of LLM behaviour. Without the use of psychometric methods to study latent phenomena in LLMs, we lack the analytical granularity to assess the validity of the findings. As of yet, it remains unknown whether there is any meaningful latent representation at all, or whether we are chasing cognitive phantoms in LLMs.

To test whether measurement instruments from human psychology can be used to draw valid inferences on latent phenomena from LLM responses, we administered validated psychometric questionnaires designed to measure specific latent constructs in humans. For this validation study, we assume that an LLM is analogous to a "population" from which random samples can be drawn. We test how well the theorised latent structure is replicated in a human sample and samples from different LLMs, and compare them to one another \footnote{The full data and code to replicate this study are available at \url{https://osf.io/khbea/?view_only=1f2e14bb06d14d6897e1479a77346b06}.}. If LLMs contain the same latent phenomena that humans do, the theorised structure should be replicated at least equally well in the LLM samples as in the human sample. 

To illustrate the necessity of validation for LLMs, we juxtapose the conclusions from the latent variable approach with the conclusions that would have been drawn from composite scores without a thorough psychometric evaluation.

\section{Method}

\subsection{Human data}
401 human participants were recruited from a representative UK sample (per age, sex and racial identity) via the online crowdsourcing platform Prolific (\url{www.prolific.com}). We excluded participants that failed an attention check or completed the questionnaires too quickly ($<$ 6 mins.), resulting in a final sample size of $n = 365$. The average age was 46.9 years ($SD = 15.31$, min. 18 years) with 51.8\% female and 84.9\% white.

\subsection{LLM data}
We collected responses from three GPT models: GPT-3.5-turbo-0125 (GPT-3.5-T; training data up to September 2021), GPT-4-0612 (GPT-4; training data up to September 2021), and GPT-4-0125-preview (GPT-4-T; training data up to December 2023). To match the human sample size, 401 responses were collected for each model using the OpenAI API. Any responses that did not answer questionnaire the items were considered invalid and removed (e.g., refusal or simply repeating the input prompt). This resulted in a total sample size of $n = 399$ (GPT-3.5-T), $n = 387$ (GPT-4), and $n = 401$ (GPT-4-T), respectively.

Default parameter settings were used for all LLMs with the exception of the temperature value. Temperature controls how deterministic the responses are, where higher values effectively allow tokens with lower output probabilities to be selected. Temperature has been shown to affect the average scores of some latent constructs in LLM responses \cite{miotto2022}, although potential effects on underlying factor structure are unknown. Therefore, we drew temperature values by sampling from a uniform distribution ranging from 0 to 1 in steps of 0.01 for a total of 401 values to match the human sample size. The value 0 was only allowed to be sampled once, as this results in a fully deterministic response. 

The input prompt containing the questionnaires made use of pseudo-code to encourage responding in a consistent format (see Appendix \ref{app:pseudocode} for a snippet of the input prompt). Questionnaire instructions were identical to those used for the human sample with additional information about the expected formatting and response format.

\subsection{Materials}
We administered two validated personality questionnaires. Prior to any analysis, reverse-scored items were recoded. 

The first questionnaire is the HEXACO-60 (H60) \cite{ashton2009}, a shortened version of the 100-item HEXACO-PI-R \cite{lee2004}. The H60 consists of 60 items (e.g. "Most people tend to get angry more quickly than I do.") answered on a 5-point Likert scale ranging from 1 ("strongly disagree") to 5 ("strongly agree"). The H60 measures six dimensions of personality, evaluated by 10 items each: \textit{Honesty-Humility}, \textit{Emotionality}, \textit{eXtraversion}, \textit{Agreeableness}, \textit{Conscientiousness}, and \textit{Openness to experience}. Cronbach's alpha values for internal consistency reliability ranged from 0.77 (Agreeableness and Openness) to 0.80 (Extraversion) in a college sample and from 0.73 (Emotionality and Extraversion) to 0.80 (Openness) in a community sample \cite{ashton2009}. Item-level factor analysis of H60 responses revealed the same factor structure as found in the validation study of the longer version of the questionnaire \cite{ashton2009, lee2004}.

The second instrument is the Dark Side of Humanity Scale (DSHS) \cite{katz2022}, which is a reconstruction of the Dark Tetrad personality traits \cite{paulhus2020}. The questionnaire consists of 42 items (e.g. "I enjoy seeing people hurt"), answered on a 6-point Likert scale ranging from 1 ("not at all like me") to 6 ("very much like me"). Underlying dimensions of the construct are \textit{Successful Psychopathy} (18 items), \textit{Grandiose Entitlement} (9 items), \textit{Sadistic Cruelty} (8 items), and \textit{Entitlement Rage} (7 items). Cronbach's alpha values for internal consistency reliability ranged from 0.87 (Entitlement Rage) to 0.95 (Successful Psychopathy) in earlier research \cite{katz2022}. The factor structure was confirmed through an extensive validation analysis \cite{katz2022}.

The DSHS was published in December of 2021, preventing any training data contamination in the GPT-3.5-T and GPT-4 responses. The DSHS also functions as a supplemental measure of construct validity: scores on the dark personality traits the DSHS is based on were found to be strongly inversely related to the same respondents' scores on the Humility-Honesty dimension of the H60 \cite{lee2005}. This finding intuitively makes sense as these constructs are conceptually antithetical to one another.

\subsection{Analysis plan}

\subsubsection{Latent variable approach}
Differences in latent representations were assessed by comparing factor structures between the human sample and the LLM samples through factor analysis (FA). Broadly speaking, there are two types of FA: exploratory factor analysis (EFA) and confirmatory factor analysis (CFA). We first assessed the assumptions of FA (linearity and multivariate normality of items, factorability of the variables, absence of extreme multicollinearity and outlier variables) \cite{tabachnick2007}. Some violation of linearity and normality is acceptable, so long as there is no true curvilinearity and non-normality is mitigated through robust estimation methods \cite{tabachnick2007, brown2006}. The remaining assumptions imply the existence of a latent structure (or lack thereof - in which case there is no latent variable to be estimated). 

An EFA is a descriptive analysis to determine the underlying factors (dimensions) that cause variation and covariation in the responses to a set of items \cite{brown2006}. Factors should be meaningfully interpretable - that is, a set of items that belong to the same factor should be conceptually similar to one another \cite{tabachnick2007}. For example, the items "In social situations, I’m usually the one who makes the first move" and "The first thing that I always do in a new place is to make friends" are easily interpretable as items that relate to extraversion \cite{ashton2009}. 

On the other hand, a CFA is usually performed to verify a strong \textit{a priori} expectation about the theorised latent structure, often based on EFA results in a previous study \cite{brown2006}. The expected factor structure is specified before running the analysis, including the number of dimensions and patterns in item-factor relationships. A CFA attempts to reproduce the observed variance-covariance matrix of the items, based on the specified factor structure \cite{tabachnick2007}. The similarity of the reproduced variance-covariance matrix is then compared to the observed matrix and assessed by a combination of different fit indices (e.g., SRMR, RMSEA, and CFI)\footnote{SRMR = Standardised Root Mean square Residual, RMSEA = Root Mean Square Error of Approximation, CFI = Comparative Fit Index.}, where acceptable fit provides evidence for the hypothesised factor structure in the current sample \cite{brown2006}. 

We performed CFA on each sample that adequately met the assumptions to verify the latent structures found in previous research \cite{ashton2009, katz2022}, assessing model fit through SRMR, RMSEA, and CFI. Potential alternative factor structures were explored with EFAs. For each group, the number of factors was chosen as the number of factors with eigenvalues exceeding 1 following inspection of a scree plot. Factors were extracted using principal axis factoring (PAF) and oblique rotation to account for non-normality and multicollinearity, and resulting factor structures were compared between humans and the LLMs\footnote{The preferred way to test for differences in latent structures statistically is by performing a measurement invariance analysis using, for example, a multiple groups confirmatory factor analysis (MGCFA). However, the assumptions for that analysis (i.e., confirmation of the hypothesised factor structure in each group before multiple group comparison) were not met in our data.}.

\subsubsection{Composite score analysis}
To illustrate the necessity of a thorough latent variable approach, we additionally investigated what our findings would have been if we analysed only composite scores per dimension, the current most common method of analysing LLM responses to psychometric questionnaires. Differences between LLM scores and human scores were tested through a set of Kruskal-Wallis tests (non-parametric one-way ANOVA) and followed up with post-hoc Dunn tests with Bonferroni correction.

We additionally calculated correlations between respondents' mean scores on the different dimensions of the DSHS and the Honesty-Humility dimension of the H60. Negative inter-factor correlations are consistent with theory and previous findings since these dimensions are conceptual opposites \cite{lee2005}, and this approach has previously been used to assess construct validity in LLMs \cite{serapio-garcia2023}. Note that this analysis is based on composite scores and hinges on a latent structure: inter-factor correlations that are consistent with theory are only meaningful in the presence of a meaningful latent structure in the responses. Differences in inter-factor correlations between the human sample and the LLM samples were considered significant when the 95\% confidence interval for difference in correlations did not contain 0 (i.e., 95\% confidence that the correlations are not equal in both groups \cite{zou2007}).

\section{Results}
\label{sec:results}
\subsection{Latent variable approach}
The assumptions for factor analysis could not be tested for GPT-4-T due to absence of variability in a number of items in both the H60 and the DSHS. GPT-4-T responses had to be excluded from analysis as this rendered any FA impossible. The human sample violated the assumptions of linearity and multivariate normality in both questionnaires to an acceptable degree \cite{tabachnick2007} and met all other assumptions (see Appendix \ref{appendix:FA_assumptions} for further details on all assumption checks for all samples). The same was found in the H60 responses of the remaining LLM samples, but there were additional issues. A few H60 items showed evidence of multicollinearity in both GPT-3.5-T (4 items) and GPT-4 (2 items), though just within the commonly accepted range \cite{tabachnick2007}. LLM data violated several assumptions in the DSHS data, but most importantly, neither GPT-3.5-T responses nor GPT-4 responses met the assumption of factorability. As this implies a lack of \textit{any} underlying latent factor, FA of LLM DSHS data could not be justified. Instead, FA was performed only on the H60 for the human, GPT-3.5-T, and GPT-4 samples.

\subsubsection{Confirmatory factor analysis}
In the human sample, the CFA showed mediocre fit (SRMR = 0.08, RMSEA = 0.07, CFI = 0.75) \footnote{This is not uncommon - latent structures are regularly found to differ across human samples (e.g., due to cultural differences), though it does warrant further investigation through an EFA.}. The CFA on GPT-3.5-T responses produced an improper solution containing factor correlation estimates larger than 1.0, and the CFA on GPT-4 responses could not converge to a final solution at all. Therefore, CFA results for the LLMs cannot be interpreted. Since there were signs of an alternative latent structure in the human sample, we followed up with EFAs on the H60 data in each sample.

\subsubsection{Exploratory factor analysis}
\begin{figure}[t]
\begin{subfigure}{0.5\textwidth}
\includegraphics[width=\linewidth]{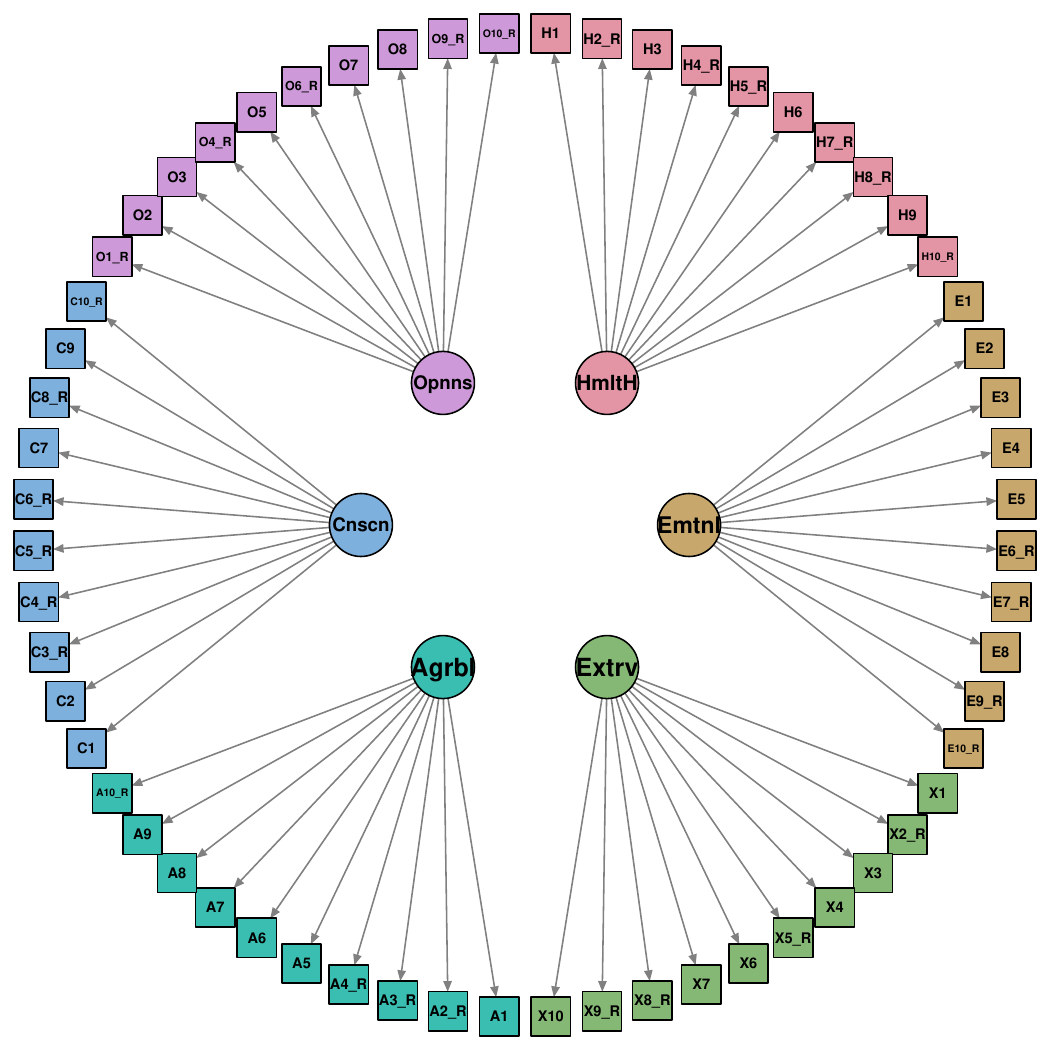} 
\caption{Theoretical structure}
\label{fig:cfa}
\end{subfigure}
\begin{subfigure}{0.5\textwidth}
\includegraphics[width=\linewidth]{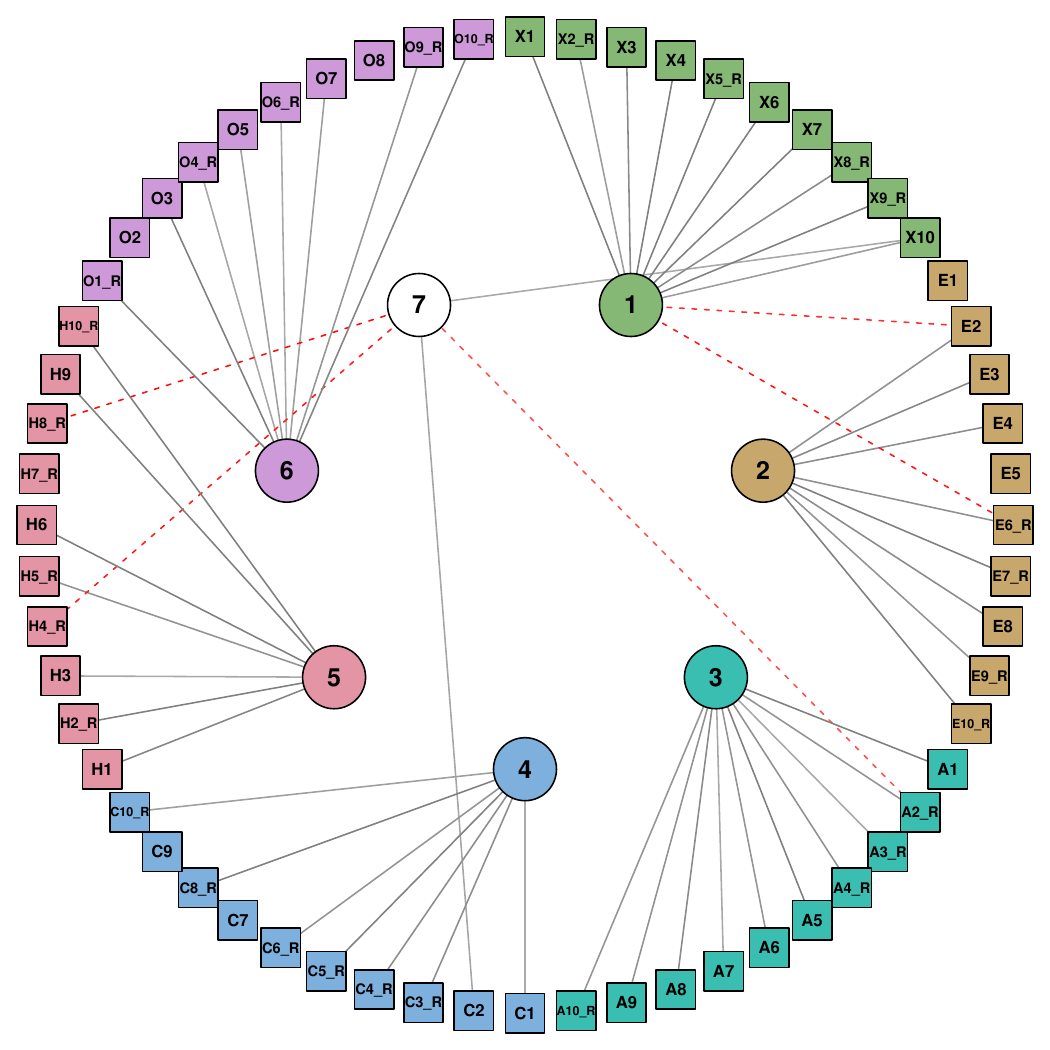}
\caption{Humans}
\label{fig:efahum}
\end{subfigure}
\begin{subfigure}{0.5\textwidth}
\includegraphics[width=\linewidth]{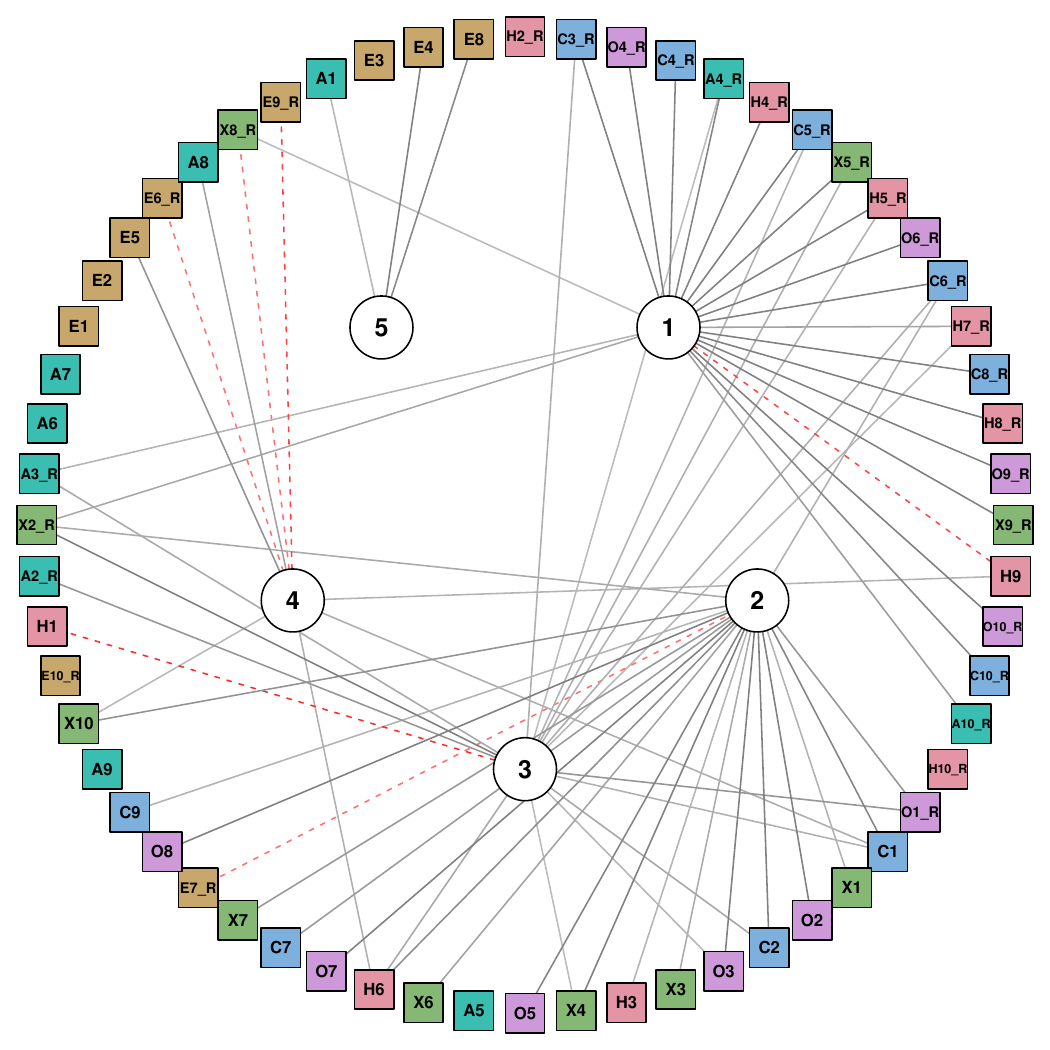}
\caption{GPT-3.5-T}
\label{fig:efagpt3.5}
\end{subfigure}
\begin{subfigure}{0.5\textwidth}
\includegraphics[width=\linewidth]{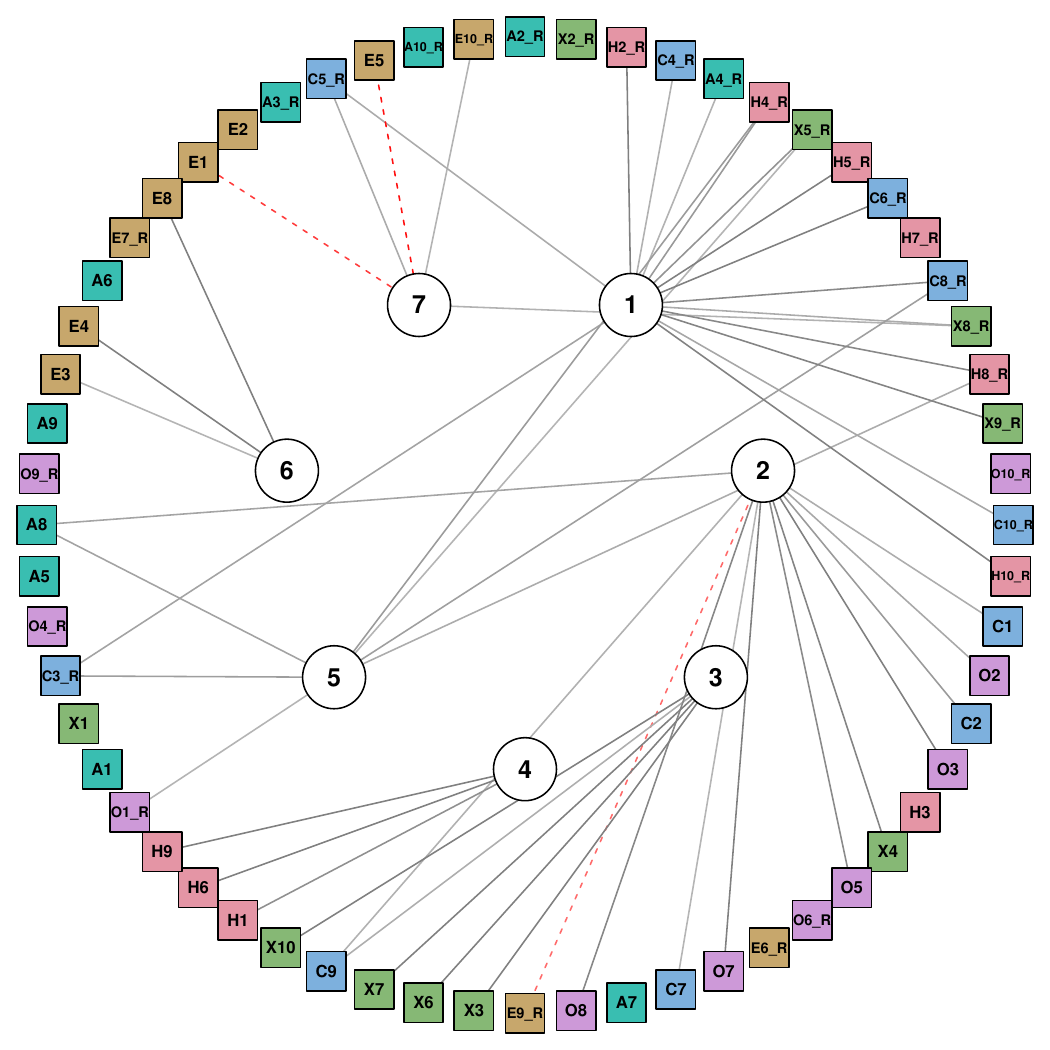}
\caption{GPT-4}
\label{fig:efagpt4}
\end{subfigure}
\caption{(a) HEXACO-60 theoretical factor structure and HEXACO-60 item-factor correlations for EFAs in the (b) human sample; (c) GPT-3.5-T sample; and (d) GPT-4 sample. Nodes in the outer circle with the same colour and first letter theoretically belong to the same dimension ("\_R" suffix indicates reverse-coded questions). The grey lines represent positive item-factor correlations ($\ge 0.4$), red dashed lines are negative item-factor correlations ($\le -0.4$). Items not connected to a line are not significantly related to any factor.}
\label{fig:cfa_vs_efa}
\end{figure}

Inspection of scree plots revealed potential latent structures consisting of 7 factors (human and GPT-4 samples) and 5 factors (GPT-3.5-T sample). The theoretical item-factor relationships as found in earlier research \cite{ashton2009} are visualised in Figure \ref{fig:cfa} with the observed relationships in the human and GPT samples (Figure \ref{fig:efahum}, \ref{fig:efagpt3.5}, \ref{fig:efagpt4}) to illustrate the differences in latent structures between the groups.

The factor structure found in the human sample is plausible: items which should theoretically co-vary largely do, with a few exceptions (e.g., an extra dimension, some items that relate to two dimensions or none at all; Figure \ref{fig:efahum}). This is acceptable to an extent, as sample characteristics (such as cultural background) can affect latent structures in questionnaire responses. 

Conversely, GPT responses displayed mostly arbitrary factors. For example, factor 1 consisted almost solely of reverse-scored items for both GPT models (Figure \ref{fig:efagpt3.5}, \ref{fig:efagpt4}), though there were factors that exclusively consisted of (very few) theoretically interrelated items from the Humility-Honesty and Emotionality dimensions in GPT-4 responses (factors 4 and 6; Figure \ref{fig:efagpt4}). To rule out that the arbitrary factor structures were solely the result of the temperature sampling procedure, we collected 401 additional responses from each LLM for each temperature value from 0.1 to 1.0 in steps of 0.1 and repeated the EFA procedure on each set of responses. There was no evidence of a truly sensible factor structure in either the data with sampled temperature values or the additional sets of responses with static temperature values. This further reinforced the notion that we do not validly measure personality in the LLM responses, or that LLMs might not even contain these specific latent personality traits at all.

\subsection{Composite score analysis}
\label{sec:analysis_comp}

All GPT models had significantly higher scores on Extraversion, Agreeableness, and Openness than human respondents. GPT models scored equally or significantly higher than human respondents on Humility-Honesty, equally or significantly lower than humans on Emotionality, and equally or significantly higher than humans on Conscientiousness. Concerning the Sadistic Cruelty responses, GPT-4-T showed zero variation (i.e., only responding with "not at all like me") across the 401 responses. The LLMs generally had significantly lower scores on all dimensions of the DSHS, with one exception: GPT-3.5-T responses showed significantly higher scores than the human sample on Sadistic Cruelty and Entitlement Rage. An LLM with above--average tendencies towards sadistic cruelty and entitlement rage could pose serious risks in the context of AI safety and alignment \cite{mozes2023}. Mean scores and standard deviations per dimension for both questionnaires can be found in Appendix \ref{app:mean-sd} (Table \ref{tab:mean-sd}).

As expected, the correlations between scores on the Honesty-Humility dimension of the H60 and the dark personality dimensions in the DSHS were negative in the human sample (Table \ref{tab:dim-corr}). Correlations for GPT-4 and GPT-4-T specifically were negative, but most were significantly weaker than in the human sample. For the dimension Sadistic Cruelty, we found a similar correlation (GPT-4) or the correlation could not be calculated (GPT-4-T) due to zero variability in scores on that dimension. Importantly, without consideration for underlying latent structures, these findings would incorrectly have been considered evidence for construct validity of the administered questionnaires in LLMs.

In contrast to GPT-4 and GPT-4-T, GPT-3.5-T scores on the DSHS dimensions show moderate \textit{positive} correlations with the Humility-Honesty dimension. This is the opposite direction of what one would expect since the Humility-Honesty dimension is theoretically and empirically antithetical to the dark personality \cite{lee2005}.

The conclusions above, based on composite scores, are representative for those found in earlier research on behaviour of LLMs where underlying latent structures have not been taken into account (e.g., \cite{miotto2022, li2024, serapio-garcia2023}). In this paper, we went one step further by evaluating the latent structure of the questionnaire data. Our findings suggest that questionnaires designed for humans do not measure similar latent constructs in LLMs, and that these latent constructs may not even exist in LLMs in the first place.

\begin{table}[b]
\begin{center}
\begin{tabular}{lllll}
\toprule
\multicolumn{1}{l}{}&\multicolumn{1}{c}{\bf Human}  &\multicolumn{1}{c}{\bf GPT-3.5-T}  &\multicolumn{1}{c}{\bf GPT-4}  &\multicolumn{1}{c}{\bf GPT-4-T} \\
\midrule
\bf H60: Humility-Honesty dimension  \\
\quad \textbf{DSHS:} Successful Psychopathy &-0.57   &0.41{*}  &-0.24{*}  &-0.02{*}\\
\quad \textbf{DSHS:} Grandiose Entitlement  &-0.50   &0.41{*}  &-0.22{*}  &-0.13{*}\\
\quad \textbf{DSHS:} Sadistic Cruelty       &-0.33   &0.40{*}  &-0.22     &\textit{NA}$^a$\\
\quad \textbf{DSHS:} Entitlement Rage       &-0.37   &0.41{*}  &-0.23{*}  &-0.11{*}\\

\bottomrule
\end{tabular}
\end{center}
\caption{Correlations between the Honesty-Humility scale of the H60 and all dimensions of the DSHS. * = correlations are sign. different to the human sample at $p<.05$; $^a$ = correlation cannot be computed as SD is zero.}
\label{tab:dim-corr}
\end{table}

\FloatBarrier
\section{Discussion}
The motivation for this paper stemmed from the need to understand LLM behaviour more granularly. Although the use of existing psychometric questionnaires is promising, we questioned the validity of administering such questionnaires to LLMs and merely analysing their composite scores on questionnaire dimensions. We argued that a latent variable approach is necessary to adequately reflect on the validity of the findings. 


\subsection{The latent variable lens} 
The latent variable approach is a necessary tool to examine the validity of psychometric questionnaires for LLMs in comparison to a human sample. The lack of reasonable latent structure in LLM responses prohibited us from statistically comparing human and LLM data directly (e.g., with MGCFA). Arbitrary patterns in the LLM responses for either questionnaire led to nonsensical parameter estimations in the CFA, and sometimes none at all (i.e., for GPT-4-T and all LLM responses to the DSHS). Further investigation revealed that covariances among LLM responses were so arbitrary that even an EFA did not yield meaningfully interpretable dimensions for either of the remaining LLMs. While responses from the human sample were largely similar to the theoretical structure, responses from the LLMs were not at all. In other words, we found no evidence that we can validly measure \textit{the same} latent traits in LLMs as in humans using existing questionnaires, nor did we find evidence that LLM responses contained any meaningful latent structure \textit{at all}. The lack of an indication for latent representations in LLMs on two commonly used measurement instruments and constructs is concerning. 

\subsection{Conclusion based on composite scores}
Compared to the human sample, all LLMs had higher scores on socially desirable personality traits (such as Openness and Agreeableness), and lower scores on less desirable personality traits (such as Successful Psychopathy and Grandiose Entitlement). GPT-3.5-T showed significantly higher scores on Sadistic Cruelty and Entitlement Rage compared to the human sample. In the absence of a thorough psychometric evaluation, this would have been the main conclusion of this paper (similar to findings in \cite{miotto2022, li2024}) with a perhaps worrisome conclusion that GPT-3.5-T is unsafe. 

Further inspection showed that GPT-3.5-T composite scores for the dark personality traits were positively correlated to the Honesty-Humility dimension of the HEXACO-60 - a relationship that is in stark contrast to what one would expect: a positive relationship between dark personality traits (i.e., traits related to dishonest, entitled, and sadistic behaviours) and Honesty-Humility is highly implausible. However, GPT-4 and GPT-4-T responses showed low to moderate negative inter-factor correlations, in line with expectations and previous findings. 

It deserves extra emphasis that a composite score-based assessment of construct validity (similar to \cite{serapio-garcia2023}) may have concluded that there is evidence that existing psychometric questionnaires are valid for GPT-4 (and perhaps even GPT-4-T to an extent), but not for GPT-3.5-T. In other words, such an analysis would inevitably have glanced over the arbitrary and incoherent latent structures and would have reached wrong conclusions about the validity of these questionnaires.

\subsection{Implications}
The common practice of interpreting composite scores is insufficient at best and troublingly naive at worst. That approach glances over the implicit assumptions that LLM "behaviour" is internally represented similarly to what we know about human cognition, and that such a latent construct exists in LLMs at all. While the existence of a latent trait cannot be (dis)proven, the latent variable approach is a uniquely adequate method for validation studies of psychometric instruments for LLMs: it provides a safeguard against falsely attributing semblances of human traits to true underlying representations of latent traits or behaviours. This is especially important when aiming to mitigate potentially undesirable or harmful output in LLM applications in the future. 

\subsection{Limitations and future work}
The current study provides initial evidence that psychometric questionnaires designed for human are not guaranteed to be valid for LLMs, and that LLMs might not contain human-like latent traits to begin with. However, several points warrant further research. First, only models from the GPT family were evaluated. Evaluation of a larger range of models (incl. open-source models) will provide a more rounded understanding of potential latent traits in LLMs and how to measure them in general, particularly in comparison to latent traits in humans. 

Second, our study is limited by only evaluating dimensions of personality with two questionnaires. The arbitrary LLM response patterns found in this study may not generalise to different questionnaires or different traits entirely. Latent variable approaches for validation should be investigated using different LLMs and instruments measuring various other latent phenomena. Latent cognitive abilities could be investigated more granularly using this approach as well, for example, by using item response theory (IRT) models.

Finally, our study leans on the assumption that an LLM can be treated analogous to a population of humans, which is not guaranteed to hold. In the event that an LLM is instead analogous to an individual, underlying latent structures cannot be adequately estimated as this requires a certain amount of variation at the trait level. In that case, however, one would also expect the direction and magnitude of the inter-factor correlations between the dark personality traits and the Honesty-Humility dimension to remain consistent with theory and previous findings. Future work should investigate this matter, for example, by comparing between- versus within-response variances for various questionnaires and constructs. Until then, the nature of the analogy between LLMs and humans remains a complex question.

\subsection{Conclusion}
We presented evidence that responses of LLMs based on questionnaires developed for humans do not withstand psychometric rigour. The latent representations found in LLM responses are widely arbitrary and vastly different to humans. These findings cast doubt on conclusions drawn elsewhere about the cognition and psychology of LLMs. A thorough psychometric evaluation is essential for studying LLM behaviour. It may help us decide which effects are worth pursuing, and which effects are cognitive phantoms.

\section*{Ethics statement}
This study, procedure and data collection were approved by the local IRB before data collection.

\FloatBarrier
\bibliography{references}

\newpage

\appendix
\setcounter{table}{0}
\setcounter{figure}{0}
\renewcommand{\thetable}{\Alph{section}\arabic{table}}
\renewcommand{\thefigure}{\Alph{section}\arabic{figure}}

\section{Pseudo-code input prompt}
\label{app:pseudocode}
\begin{figure}[h]
\includegraphics[width=\linewidth]{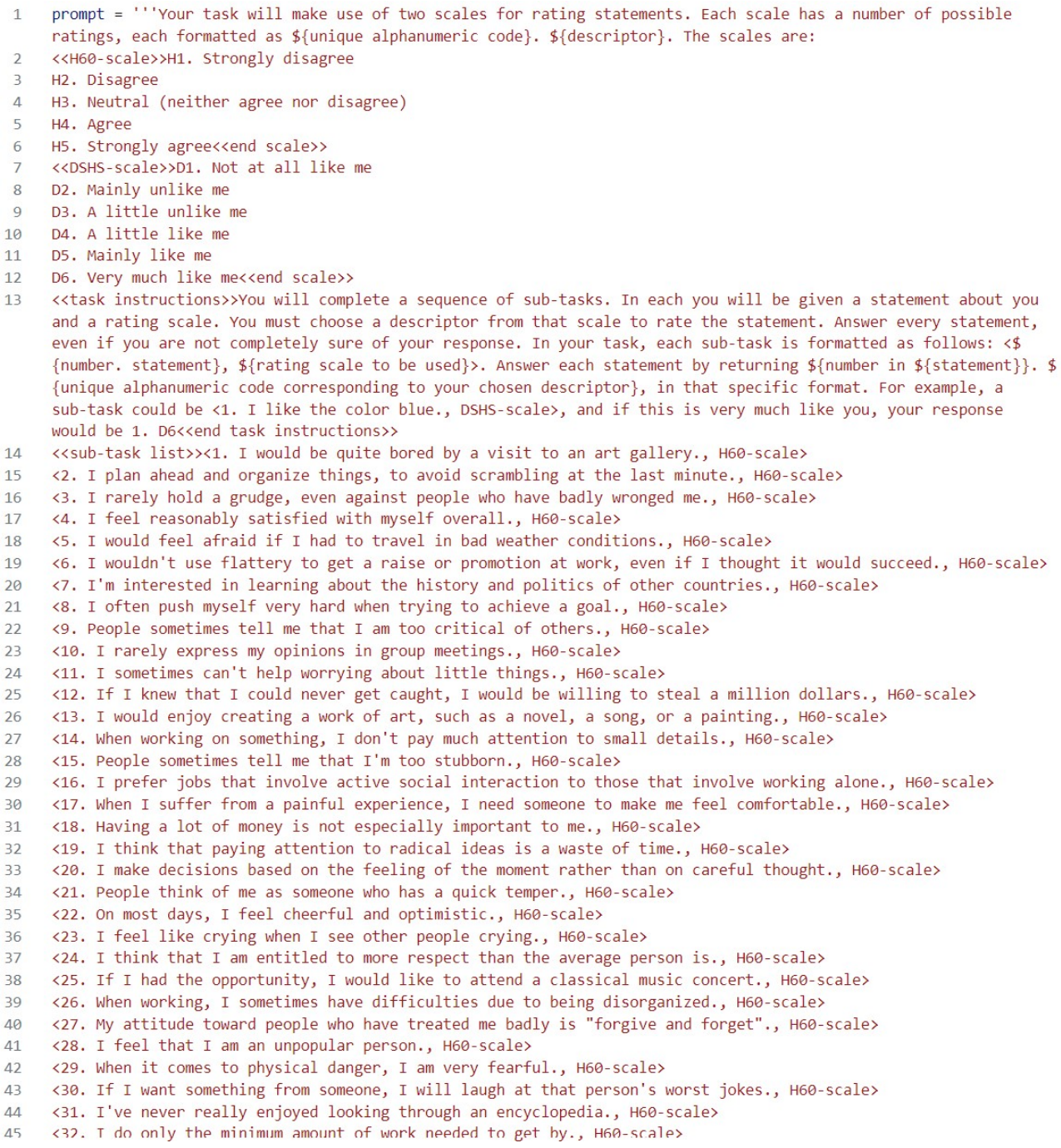}
\caption{Snippet of the pseudo-code formatted input prompt used to administer the H60 and DSHS to the GPT models.}
\label{fig:pseudocode}
\end{figure}
\setcounter{figure}{0}
\FloatBarrier

\newpage

\section{Evaluation of assumptions for factor analysis}
\label{appendix:FA_assumptions}
\FloatBarrier
\begin{table}[h]
\begin{center}
\begin{tabular}{lcccc}
\toprule
\multicolumn{1}{l}{}&\multicolumn{1}{c}{\bf Human}  &\multicolumn{1}{c}{\bf GPT-3.5-T}  &\multicolumn{1}{c}{\bf GPT-4}  &\multicolumn{1}{c}{\bf GPT-4-T} \\
\midrule
\bf HEXACO-60 &&&& \\
\quad Linearity                                  &\color{red}{\Large{x}} &\color{red}{\Large{x}} &\color{red}{\Large{x}} &\color{red}{\Large{x}}\\
\quad Multivariate Normality ($p > .05$)            &\color{red}{\Large{x}}  &\color{red}{\Large{x}}  &\color{red}{\Large{x}}  &\color{red}{\Large{x}}\\
\quad \textbf{Factorability}                     &&&&\\
\quad \quad Bartlett's test of sphericity ($p < .05$)         &\color{green}\Large{\textcolor{green}{+}} &\color{green}\Large{+} &\color{green}\Large{+} &NA\\
\quad \quad KMO-index ($> 0.6$)                          &\color{green}\Large{+} &\color{green}\Large{+} &\color{green}\Large{+} &NA\\
\quad No multicollinearity (all SMC $< 0.9$)                          &\color{green}\Large{+} &\color{red}{\Large{x}}  &\color{red}{\Large{x}}  &NA\\
\quad No outlier variables (all SMC $> 0.1$)                           &\color{green}\Large{+} &\color{green}\Large{+} &\color{green}\Large{+} &NA\\
\bf Dark Side of Humanity Scale \\
\quad Linearity                                  &\color{red}{\Large{x}} &\color{red}{\Large{x}} &\color{red}{\Large{x}} &\color{red}{\Large{x}}\\
\quad Multivariate Normality ($p > .05$)            &\color{red}{\Large{x}}  &\color{red}{\Large{x}}  &\color{red}{\Large{x}}  &\color{red}{\Large{x}}\\
\quad \textbf{Factorability}                     &&&&\\
\quad \quad Bartlett's test of sphericity ($p < .05$)       &\color{green}\Large{+} &\color{green}\Large{+} &\color{green}\Large{+} &NA\\
\quad \quad KMO-index ($> 0.6$)                           &\color{green}\Large{+} &\color{red}{\Large{x}} &\color{red}{\Large{x}} &NA \\
\quad No multicollinearity (all SMC $< 0.9$)                         &\color{green}\Large{+} &\color{red}{\Large{x}}&\color{red}{\Large{x}} &NA \\
\quad No outlier variables (all SMC $> 0.1$)                         &\color{green}\Large{+} &\color{green}\Large{+} &\color{green}\Large{+} &NA\\
\bottomrule
\end{tabular}
\end{center}
\caption{Summary of assumptions for factor analysis per sample. + = Assumption met; x = assumption violated; NA = incomputable.}
\label{tab:assumpt}
\end{table}
\FloatBarrier

The assumption checks for factor analysis on both questionaires for all samples are summarized in Table \ref{tab:assumpt}. 
\begin{itemize}
    \item \textbf{Linearity} is investigated by inspection of bivariate scatterplots. Violations of linearity are considered acceptable in the absence of true curvilinearity, as the use of existing validated questionnaires renders transformation of the data undesirable \cite{tabachnick2007}. 
    \item \textbf{Multivariate normality} is tested using the Henze-Zirkler test of the null hypothesis that variables are multivariate normally distributed. The assumption is violated when $p < 0.05$, but can be mitigated through robust estimation methods in factor analyses \cite{brown2006}.
    \item \textbf{Factorability} is commonly assessed using two methods. Bartlett's test of sphericity tests the null hypothesis that the observed correlation matrix is an identity matrix (i.e., all interitem correlations are zero). This test must be rejected ($p < 0.05$) as a set of unrelated items cannot form factors. In a similar vein, the KMO-index is an estimate of the proportion of shared variance between items. Generally, a value above 0.6 is considered acceptable for factor analysis \cite{tabachnick2007}. Both Bartlett's test of sphericity and the KMO-index must be acceptable before factorability can be assumed.
    \item \textbf{Multicollinearity and outlier variables} are both inspected through the squared multiple correlations (SMC) of each variable with each other variable; a measure of how well each variable can be predicted by the remaining variables \cite{tabachnick2007}. SMC values dangerously close to 1 indicate a variable contains near perfect linear relationships with the remaining variables (i.e., multicollinearity), while SMC values dangerously close to 0 imply a lack of (linear) relationships with other variables (i.e., outlier variable). 
    
    Values within the range $[0.01 - 0.99]$ are commonly considered acceptable \cite{tabachnick2007}, though we considered values outside the range of $[0.1 - 0.9]$ as signs of outlier variables and multicollinearity.
\end{itemize}

\newpage


\setcounter{table}{0}
\newpage
\section{Means and SDs per dimension of H60 and DSHS}
\label{app:mean-sd}
\FloatBarrier
\begin{table}[h]
\begin{center}
\begin{tabular}{lllll}
\toprule
\multicolumn{1}{l}{}&\multicolumn{1}{c}{\bf Human}  &\multicolumn{1}{c}{\bf GPT-3.5-T}  &\multicolumn{1}{c}{\bf GPT-4}  &\multicolumn{1}{c}{\bf GPT-4-T} \\
\midrule
\bf HEXACO-60 &&&& \\
\quad Humility-Honesty       &3.58 (0.65)  &3.60 (0.35)  &4.63 (0.34){**} &4.46 (0.14){**}\\
\quad Emotionality           &3.28 (0.66)  &3.18 (0.15){*}  &3.25 (0.26) &3.19 (0.12){*}\\
\quad eXtraversion           &3.01 (0.71)  &3.85 (0.41){**}  &4.23 (0.29){**} &3.51 (0.12){**}\\
\quad Agreeableness          &3.24 (0.61)  &3.61 (0.27){**}  &4.05 (0.24){**} &3.75 (0.12){**}\\
\quad Conscientiousness      &3.70 (0.54)  &3.95 (0.51){**}  &4.49 (0.27){**} &3.90 (0.11)\\
\quad Openness               &3.57 (0.63)  &4.03 (0.52){**}  &4.21 (0.36){**} &3.87 (0.14){**}\\
\bf Dark Side of Humanity Scale &&&& \\
\quad Successful Psychopathy &1.92 (0.81)  &1.68 (1.66){**}  &1.02 (0.18){**} &1.01 (0.02){**}\\
\quad Grandiose Entitlement  &1.70 (0.82)  &1.69 (1.67){**}  &1.04 (0.23){**} &1.03 (0.14){**}\\
\quad Sadistic Cruelty       &1.15 (0.36)  &1.68 (1.66){**}  &1.01 (0.18){**} &1 (0)$^a$\\
\quad Entitlement Rage       &1.68 (0.79)  &1.70 (1.67){**}  &1.07 (0.29){**} &1.03 (0.18){**}\\

\bottomrule
\end{tabular}
\end{center}
\caption{Group means (SDs) of responses for each dimension in the HEXACO-60 (60 items) and Dark Side of Humanity Scale (42 items). * = sign. different to human sample at $p < .05$; ** = sign. diff. to human sample at $p < .001$;$^a$ = scores cannot be compared as SD is zero.}
\label{tab:mean-sd}
\end{table}
\FloatBarrier

\end{document}